\documentclass[journal]{IEEEtran}
\usepackage{hyperref}
\usepackage{cite}
\usepackage[shortlabels]{enumitem}
\usepackage{graphicx}
\usepackage{multirow}
\usepackage{amsmath}
\usepackage{makecell}
\usepackage{footnote}
\makesavenoteenv{tabular}
\usepackage{amssymb}
\usepackage{hyperref}
\DeclareMathOperator*{\argmax}{arg\,max}

\begin{document}

\title{A Fair and Efficient Hybrid Federated Learning Framework based on XGBoost \\ for Distributed Power Prediction}

\author{\IEEEauthorblockN{Haizhou Liu, Xuan Zhang*, Xinwei Shen* and
Hongbin Sun,~\IEEEmembership{Fellow,~IEEE}}

\thanks{Haizhou Liu, Xuan Zhang and Xinwei Shen are affiliated with Tsinghua-Berkeley Shenzhen Institute, Tsinghua Shenzhen International Graduate School, Tsinghua University, 518055, Shenzhen, China. Hongbin Sun is affiliated with the Department of Electrical Engineering, Tsinghua University, 100084, Beijing, China. (Corresponding Author: Xuan Zhang, \href{mailto:xuanzhang@sz.tsinghua.edu.cn}{xuanzhang@sz.tsinghua.edu.cn}; Xinwei Shen, \href{mailto:sxw.tbsi@sz.tsinghua.edu.cn}{sxw.tbsi@sz.tsinghua.edu.cn}.)} }

%\markboth{Manuscript Submitted to IEEE Transactions}%
%{Manuscript Submitted to IEEE Transactions}

\IEEEtitleabstractindextext{%
\begin{abstract}
In a modern power system, real-time data on power generation/consumption and its relevant features are stored in various distributed parties, including household meters, transformer stations and external organizations. To fully exploit the underlying patterns of these distributed data for accurate power prediction, federated learning is needed as a collaborative but privacy-preserving training scheme. However, current federated learning frameworks are polarized towards addressing either the horizontal or vertical separation of data, and tend to overlook the case where both are present. Furthermore, in mainstream horizontal federated learning frameworks, only artificial neural networks are employed to learn the data patterns, which are considered less accurate and interpretable compared to tree-based models on tabular datasets. To this end, we propose a hybrid federated learning framework based on XGBoost, for distributed power prediction from real-time external features. In addition to introducing boosted trees to improve accuracy and interpretability, we combine horizontal and vertical federated learning, to address the scenario where features are scattered in local heterogeneous parties and samples are scattered in various local districts. Moreover, we design a dynamic task allocation scheme such that each party gets a fair share of information, and the computing power of each party can be fully leveraged to boost training efficiency. A follow-up case study is presented to justify the necessity of adopting the proposed framework. The advantages of the proposed framework in fairness, efficiency and accuracy performance are also confirmed.

\end{abstract}

}

\maketitle
\IEEEdisplaynontitleabstractindextext
\IEEEpeerreviewmaketitle

\section{Introduction}
\label{Section-I}
\IEEEPARstart{P}{ower} prediction has become an increasingly important topic in the modern power system. An accurate and efficient prediction of the generated/consumed power not only enables system operators of different levels to make optimal planning and operation decisions \cite{Importance-Operator1,Importance-Operator2,Importance-Operator3}, but also enlightens electricity and ancillary service retailers on customizing their prices and services \cite{Importance-Market1,Importance-Market2}. The prevalence of machine learning further empowers power prediction with universal and high-performance data-mining algorithms. In the past decades, various literatures have employed machine learning to address power prediction, from statistical regression \cite{Linear-Regression} and time series analysis \cite{Time-Series}, to the more recent family of neural networks \cite{ANN,ANN2,ANN3} and decision trees \cite{Tree1,Tree2}, using data collected from multiple sources of power generation or consumption.

One problem inherent in power prediction is the distributed nature of these data sources. For example, district-level power consumption data are generated inside numerous smart meters, and subsequently aggregated by hundreds of transformer stations across the city. Features relevant to power prediction (e.g., population mobility data) are also held by external parties in a distributed fashion. A centralized machine learning framework would require all these distributed parties to transfer their data to a central server for training, which would jeopardize data security and lead to high communicational burden during the data transfer. 

Fortunately, federated learning, proposed by Google in 2016 \cite{Google}, provides a viable solution framework that learns a shared model by aggregating locally-computed model updates. In this framework, the proprietary datasets of each party remain immobilized, which protects both data privacy and training efficiency. Federated learning was then extensively investigated to address heterogeneity, security among other issues under various scenarios, during which two major branches emerged: (a) horizontal federated learning, which involved collaboration of multiple homogeneous parties with the same feature set but a different sample set; (b) vertical federated learning, where the collaborating parties were heterogeneous, holding the same sample set but with complementary features.

In the past three years, federated learning has started to garner interest in various aspects of the power system. Saputra \emph{et al.} \cite{EV} designed a federated energy demand prediction framework for electric vehicle networks.  Wen \emph{et al.} proposed FedDetect \cite{FedDetect} to identify energy theft behavior in the smart grid. Wang \emph{et al.} introduced federated learning into electricity consumer characteristics identification \cite{Smart-Meter-ID}. In terms of power prediction, Fekri \emph{et al.} proposed a distributed load forecasting scheme based on federated learning using recurrent neural networks \cite{Fed-Load1}, and Gholizadeh \emph{et al.} combined federated learning with hyperparameter-based clustering for electrical load forecasting \cite{Fed-Load2}. Our previous work \cite{Our-Work} systematically explored both horizontal and vertical federated learning on a power consumption dataset of 64 transformer stations across a southern China city, while introducing population mobility as external features to improve prediction accuracy.  

However, since its foundation, federated learning has been polarized towards either the horizontal or vertical framework, while neglecting the case where both are present. As is often the case, though, that features can be scattered in local heterogeneous parties while samples are scattered in different local districts, resulting in a hybrid case where data are separated both vertically and horizontally.

It is also noticeable that artificial neural networks are the mainstream machine learning algorithm built in federated learning frameworks, particularly horizontal federated learning, including the most renowned FedSGD and FedAvg frameworks \cite{FedAvg}. This is mostly attributable to the fact that neural network model updates can be properly ``averaged" by the central server in these traditional frameworks, resulting in an aggregated model as if all data were centralized for training. On the other hand, averaging the updates of locally-trained models makes little sense for tree-based models, especially for the most advanced Extreme Gradient Boosting (XGBoost) \cite{XGBoost} and LightGBM \cite{LightGBM} algorithms. There is empirical evidence, though, that tree-based models generally outperform neural networks in terms of prediction accuracy, especially on tabular datasets \cite{Tabular}. Also, neural networks are known to be lacking in interpretability unlike decision trees. In a scenario where we wish to predict power with real-time external features such as demographics while maintaining an interpretable model, tree-based algorithms would be highly preferred.

Recently, a handful of research papers on horizontal federated learning proposed alternative aggregation schemes for tree-based architectures. Instead of averaging the locally-trained model updates in each round, these schemes aggregate gradient statistics of the samples in each party, after which a central server starts a round of training (i.e., node splitting) on behalf of all parties. Liu \emph{et al.} proposed FEDXGB for mobile crowdsensing \cite{FEDXGB}, in which the central server was forced to perform aggregation before decryption in the presence of a secret sharing algorithm. Li \emph {et al.} proposed a serverless framework \cite{Fed-similarity} where each party boosted a number of trees in turn, by exploiting similarity information based on locality-sensitive hashing.  Tian \emph{et al.} proposed a horizontal and serverless FederBoost framework for private federated learning of gradient boosting decision trees \cite{FederBoost}, in which all the uploaded gradients were transferred to the active party for node splitting, and only a lightweight secure aggregation scheme was required. The essence of gradient aggregation in these frameworks, if properly combined with an XGBoost-based vertical framework such as SecureBoost \cite{SecureBoost}, can result in a hybrid federated learning framework entirely based on XGBoost, to address the case where data are both horizontally and vertically separated.

Finally, to maximally protect the privacy of each participant, a federated learning scheme independent of central servers, which are often assumed semi-honest, is pursued. Nonetheless, in a serverless tree-based federated learning framework, an ``active party" is often responsible for the training of an entire tree, during which other parties have to transfer all gradients to that party and wait passively for the node splitting results. This poses additional fairness and efficiency threats to the framework, both of which are frequently discussed topics \cite{Fair1,Fair2,Efficiency1,Efficiency2} in federated learning:

\begin{itemize}
\item Fairness. In a decision tree, the active party learns the gradient information of every participating party. Even if the gradients are properly aggregated before transferral, this gives the active party an unfair advantage at inferring other parties' information. This compromises information fairness.

\item Efficiency. In the node-splitting stage of every node, only one active party performs the calculation task. The natural advantage of parallel processing in distributed learning is therefore left aside, leading to computational inefficiency.

\end{itemize}

The problems of fairness and efficiency could have been alleviated, though, by means of a dynamic task allocation scheme during the construction and training of each tree. 

To address the aforementioned problems, we hereby propose a fair and efficient hybrid federated learning framework for the distributed power prediction of a city-level distribution network. The contribution of this paper is threefold:

\begin{enumerate}[leftmargin=*]
\item We propose a hybrid tree-based federated learning framework based on XGBoost. The hybrid nature of the framework can address both the horizontal and vertical separation of data, and the introduction of XGBoost improves prediction accuracy on tabular datasets. This is important to today's power systems as it effectively addresses the distributed nature of data in the multi-level grid.

\item We propose a dynamic task allocation scheme for the proposed framework, in which the node-splitting tasks are flexibly assigned to each party depending on their current working status. We theoretically confirm its superiority of fairness and efficiency under the ideal assumption of computational uniformity. We also use the case study to simulate the case where there is a statistical difference in the time consumption of each party. This enables the distributed parties to make contributions to pattern mining and thereby make timely decisions to smart grid planning and operation in a coordinated manner.

\item We implement the proposed framework on a real-life scenario -- the prediction for distribution networks in a city of southern China. We justify the necessity of implementing the proposed hybrid framework in our problem setup. Through XGBoost prediction and simulation of the task allocation process, we also confirm the advantages of the proposed framework in fairness, efficiency and prediction accuracy.

\end{enumerate}

The rest of the paper is structured as follows. \hyperref[Section-II]{Section II} describes the real-life problem setup of power prediction scenarios, which calls for the implementation of a hybrid federated learning framework. \hyperref[Section-III]{Section III} introduces the proposed framework. \hyperref[Section-IV]{Section IV} evaluates the performance of the framework theoretically. \hyperref[Section-V]{Section V} provides a detailed case study on the problem setup, confirming the effectiveness of the framework. \hyperref[Section-VI]{Section VI} concludes the article. 

\section{Problem Statement}
\label{Section-II}
\begin{figure}
\centering
\includegraphics[scale=0.3]{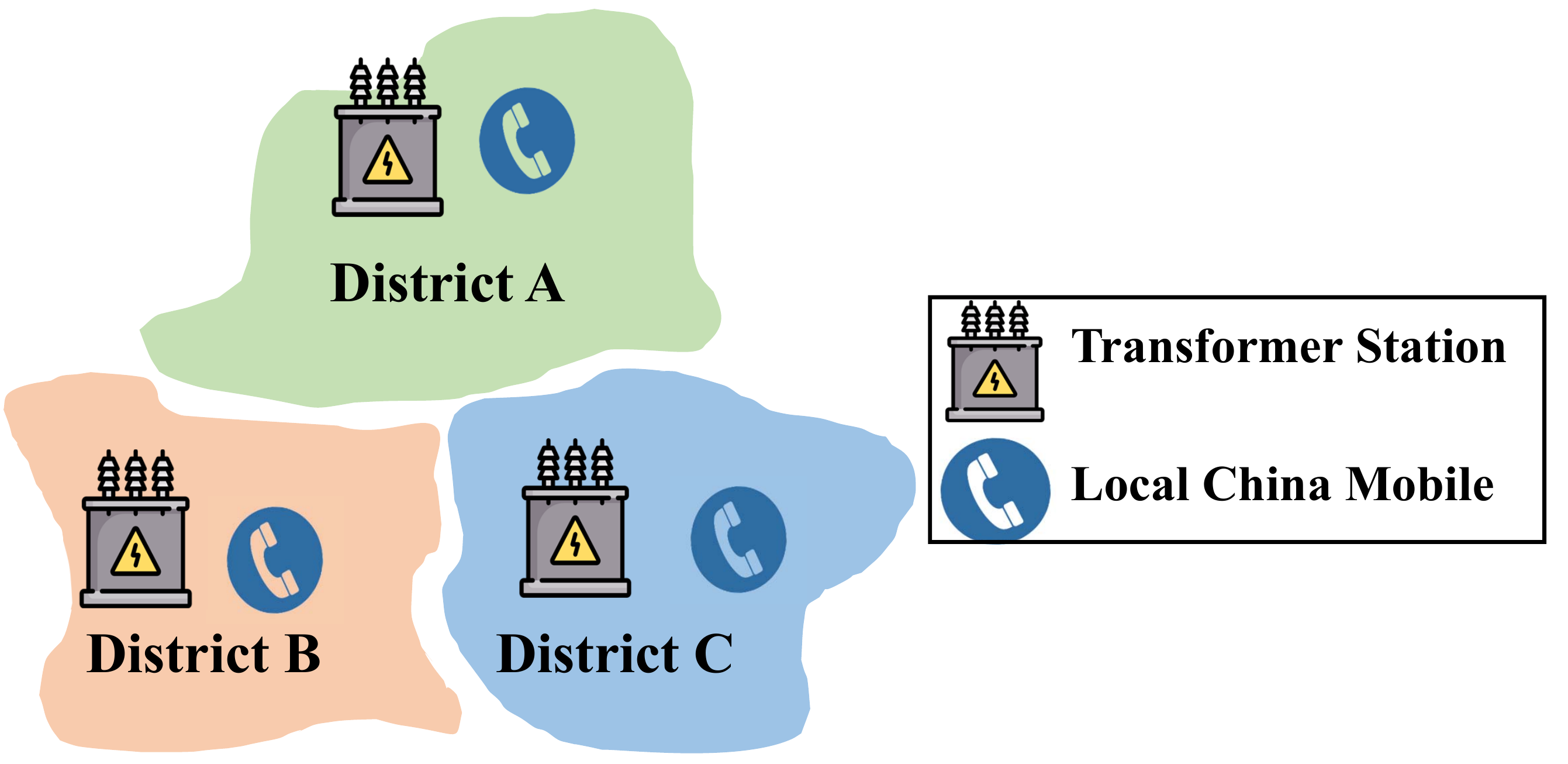}
\caption{An illustration of how the data in the problem setup are stored in distributed parties.}
\label{Illustration}
\end{figure}

This study is based on a Smart City project carried out in a southern city of China, similar to the problem setup of our previous work \cite{Our-Work}. In the project, the power consumption data are stored in 64 district-level transformer stations across the city, which are directly coordinated by the Power Supply Bureau. Weather information, scraped from the TimeAndDate website \cite{Time-And-Date}, are also treated as real-time internal features of the local transformer stations. In addition, China Mobile, China's largest telecommunication company, signed an agreement in 2019 to supplement the project with district-level demographic information as external features. These features include real-time district portraits (headcount, age, gender, etc.) and the district's Pearson correlation factor with typical zones (industrial, commercial, etc.). All features available from the three datasets are summarized in \hyperref[dataset]{\textbf{Table I}}. The data are stored in a distributed fashion separated by districts and parties, as sketched in \hyperref[Illustration]{\textbf{Figure 1}}.

\begin{table*}[ht]
\label{dataset}
\begin{center}

\caption{Description of all three datasets used for the problem setup.}
\renewcommand\arraystretch{1.3}
\begin{tabular}{c  c  c  c}
\hline
\hline
& \textbf{Power Consumption Dataset} & \textbf{Weather Dataset} &\ \textbf{Mobility Dataset}\\

\hline
\textbf{Data Source} & \makecell{Transformer Stations of \\ Power Supply Bureau} & TimeAndDate.com  * & China Mobile \\
\hline
\textbf{Resolution} & 5 minutes & 1 hour & 1 hour \\

\hline
\textbf{Features} & \makecell{Power Consumption (\emph{Label}) \\Hour} & \makecell{Temperature,\\ Wind speed,\\ Humidity,\\ Barometer} & \makecell{District Headcount,\\ Gender Statistic, \\ Age Statistic (5 bins),\\ Wage Statistic (3 bins), \\Correlation with Typical Zones}\\

\hline
\hline 
 
\end{tabular}

 {\ \\ * The weather features from TimeAndDate.com are treated as internal features of the transformer stations. \\ All demographic data are inferred by individual users' log info in each district. \par}
 
\end{center}

\end{table*}

The goal of this project is to train a collaborative machine learning model $f^*$, using the transformer station $C_m$'s weather profiles $\mathcal{D}_{Cm}$ and China Mobile $D_m$'s demographic profiles $\mathcal{D}_{Dm}$ in districts $m=1,2, \cdots, M$, to predict the districts' actual power consumption $P_m$:

\begin{equation}
\text{min} \quad ||P_m-f^*(\mathcal{D}_{Cm},\mathcal{D}_{Dm})||, \quad m =1, 2, \cdots, M 
\end{equation} 

\noindent where $||\cdot||$ represents a metric for evaluating the prediction error, such as the Mean Squared Error (MSE). The learned model $f^*$ is a global function and does not differentiate among districts. Note that as opposed to typical power forecasting methods, historical power consumption records are excluded from the input feature set, in order to better interpret how power consumption is related to real-time demographic and weather profiles from the as-trained model $f^*$. This ensures that in terms of abrupt changes in these profiles, $f^*$ still remains its validity. 

Three observations can be made on this problem setup:

\begin{enumerate}[1)]
\item The input dataset is ``tabular" in nature, where each record/sample contains a mixed combination of weather features, demographic features and power consumption labels. Neural networks are known to be empirically incompetent in learning such tabular datasets, due to a lack of locality, feature heterogeneity and mixed feature types \cite{Tabular}. Decision trees, on the contrary, can effectively extract the patterns in such tabular data.

\item The records are stored in a distributed fashion by various homogeneous and heterogeneous parties. A centralized machine learning scheme where data are collected by a central server before training would compromise data privacy. Federated learning can address this issue by training locally on each device, without direct data sharing.

\item While the dataset is vertically separated into China Mobile and the Power Supply Bureau by features, it is also horizontally separated into different local districts by records. This means that neither a horizontal or vertical federated learning framework is sufficient for this problem. A hybrid framework is needed to address both types of data separation. 

\end{enumerate}

In \hyperref[Section-III]{Section III}, we will show that our proposed framework, by incorporating both horizontal and vertical federated learning with an XGBoost-based learning architecture, can appropriately address all these inherent problems. 

\section{Proposed Framework}
\label{Section-III}

In this section, we first introduce the preliminaries of XGBoost, which we will employ as our bottom-level learning architecture. Then, the proposed hybrid federated learning framework is outlined, which is built upon XGBoost as well as state-of-the-art horizontal and vertical federated learning schemes. Finally, the dynamic task allocation scheme is introduced as to improve the prediction performance.

\subsection{Preliminaries: Bottom-Level XGBoost Architecture}
The bottom-level XGBoost algorithm learns the inner patterns of data by consecutively constructing a series of gradient-boosted trees $\mathcal{T}_1,\mathcal{T}_2, \cdots, \mathcal{T}_S$. Each tree aims to predict the residual prediction error generated by previous trees. The final prediction output of XGBoost is therefore the summation of all intermediate results yielded from these trees: 

\begin{equation}
f(\mathbf{x})=\mathcal{T}_1(\mathbf{x})+ \mathcal{T}_2(\mathbf{x})+\cdots +\mathcal{T}_S(\mathbf{x})
\end{equation}

\noindent where $\mathbf{x}$ denotes the input feature vector of a sample, and $f(\cdot)$ denotes the full XGBoost model aiming to predict its label. Each gradient-boosted tree is designed as a binary decision tree: to obtain the output of a sample on a particular tree, the sample starts from the head node of the tree, and falls to either the left or right child node, based on whether a certain criterion is met for a feature of the sample. This continues until the sample reaches a leaf node, which contains the expected output of the sample (also known as leaf values).

Training is required on the XGBoost model in order to determine the best model parameters, including the node-splitting criterion and leaf values. Constructing each tree $\mathcal{T}$ while determining its optimal parameters is also known as training an epoch. XGBoost trains each epoch by first collecting the gradients of the training samples belonging to the target node. Then, for non-leaf nodes, XGBoost iterates over all samples and features to determine the best node-splitting criterion, namely the feature and threshold value that can induce the largest information gain: 

\begin{equation}
\label{non-leaf}
\begin{cases}
& \text{Gain}(k,v) = {\frac{G_l^2}{H_l+\lambda}+\frac{G_r^2}{H_r+\lambda}-\frac{G^2}{H+\lambda}} \\
& k^*,v^*=\argmax_{k,v}\{\text{Gain}(k,v)\}
\end{cases}
\end{equation}

\noindent where $k,v$ are the candidate features and thresholds for node-splitting, and $k^*,v^*$ denote their optimal values. $G_l$ and $H_l$ denote the sum of first-order and second-order gradients of the samples falling onto the current node with feature $k$ smaller  than the threshold $v$. $G_r$ and $H_r$ denote the sum of first-order and second-order gradients of the samples falling onto the current node with feature $k$ greater than the threshold $v$. $G=G_l+G_r, H=H_l+H_r$. $\lambda$ is the regularization parameter.

For leaf-nodes, XGBoost calculates the leaf values $w^*$ by:

\begin{equation}
\label{leaf}
w^*=-\frac{\eta G}{H+\lambda}
\end{equation}

\noindent where $\eta$ denotes the learning rate. 

In addition, based on the observation that the training in (\ref{non-leaf}) and (\ref{leaf}) only requires the summation statistics of the gradients, XGBoost proposes an approximate split-finding algorithm, in which we first categorize samples into bins according to their value on each feature, followed by aggregating the gradients of samples in each bin. The candidate splits are then set as the thresholds between adjacent bins (instead of between adjacent samples in the original algorithm), which significantly reduces the number of node-splitting candidates to inspect. Finally, the gradient statistics $G_l$, $G_r$ and $G$ are calculated by summing up the gradient summations of the corresponding bins.  The core algorithm for binning, node splitting and leaf value calculation is outlined in \hyperref[Framework-1]{\textbf{Framework 1}} \cite{XGBoost}.

\begin{table}[ht]
\label{Framework-1}
\begin{tabular}{l  l}

\hline
\hline
\multicolumn{2}{l}{\textbf{Framework 1: XGBoost Training (via Approximate Finding)}} \\
\hline
\multicolumn{2}{l}{\textbf{Additional Notations:}} \\
\multicolumn{2}{l}{ $I$, the sample space of the target node} \\
\multicolumn{2}{l}{$K$, Number of features; $V$, Number of bins} \\
\multicolumn{2}{l}{$x_{i,k}$, Value of the $k$-th feature of sample $i$} \\

\ \\

\multicolumn{2}{l}{\textbf{Binning and Gradient Statistics Aggregation} \emph{for every tree}.} \\
\textbf{1} & \textbf{For} $i$ in $I$: \\
\textbf{2} & \quad Calculate $g_i$ and $h_i$, the first- and second-order gradient\\
\textbf{3} & \textbf{End For} \\
\textbf{4} & \textbf{For} feature $k = 1, 2, \cdots, K$: \\

\textbf{5} & \quad Determine the edge of bins for feature $k$: $S_k=\{s_{k0}, \cdots, s_{kV}\}$\\
\textbf{6} & \quad \textbf{For} $v = 1, 2, \cdots, V:$ \\
\textbf{7} & \quad \quad $G_{k,v}=\sum_{i \in \{i| s_{k,v-1} < x_{i,k} \leq s_{k,v}\}} g_i$\\
\textbf{8} & \quad \quad $H_{k,v}=\sum_{i \in \{i| s_{k,v-1} < x_{i,k} \leq s_{k,v}\}} h_i$\\
\textbf{9} & \quad \textbf{End For}  \\
\textbf{10} & \textbf{End For}  \\

\ \\
\multicolumn{2}{l}{\textbf{Node-Splitting} \emph{for every non-leaf node of a tree}.} \\

\textbf{1} & $G=\sum_{k}G_{k,1}$, $H=\sum_{k}H_{k,1}$ \\
\textbf{2} & \textbf{For} feature $k = 1, 2, \cdots, K$: \\
\textbf{3} & \quad $G_l=0$, $H_l=0$ \\
\textbf{4} & \quad \textbf{For} bin $v = 1, 2, \cdots, V$: \\
\textbf{5} & \quad \quad $G_l=G_l+G_{kv}$, $H_l=H_l+H_{kv}$ \\
\textbf{6} & \quad \quad $G_r=G-G_l$, $H_r=H-H_l$ \\
\textbf{7} & \quad \quad $\text{Gain}(k,v)=\frac{G_l^2}{H_l+\lambda}+\frac{G_r^2}{H_r+\lambda}-\frac{G^2}{H+\lambda}$  \\
\textbf{8} & \quad \textbf{End For}  \\
\textbf{9} & \textbf{End For}  \\
\textbf{10} & $k^*$, $v^*=\argmax_{k,v}\text{Gain}(k,v)$ \\
\textbf{11} & The splitting criterion is set to a threshold of $s_{k^*v^*}$ on feature $k$ \\

\ \\
\multicolumn{2}{l}{\textbf{Leaf Value Calculation} \emph{for every leaf node of a tree}.} \\

\textbf{1} & $G=\sum_{k}G_{k,1}$, $H=\sum_{k}H_{k,1}$ \\

\textbf{2} & Set the leaf value to $\omega = -\frac{\eta G}{H+\lambda}$ \\

\hline
\hline 

\end{tabular}
\end{table}

The as-trained XGBoost model is by nature a recursive if-else structure, in which a sample traverses from the head node to a leaf node based on true/false statements of a particular feature value. XGBoost is therefore more interpretable than neural networks, the latter of which contains countless nonlinear feature combinations in neurons, making it hard to explain. XGBoost is also known to be a more powerful model with higher prediction accuracy on tabular datasets. This is one of the reasons that we adopt XGBoost as our bottom-level learning algorithm in the first place.

Another advantage of XGBoost is that the approximate finding algorithm, by reducing the number of node-splitting candidates with the binning strategy, also alleviates considerable communicational and computational burden in the transfer of parameters of the upcoming federated learning framework.

\subsection{Hybrid Federated Learning Based on XGBoost}
Federated learning constructs a secure and collaborative learning framework on top of machine learning algorithms, so that all parties can jointly train a unified prediction model, without having to share any local data. Horizontal and vertical federated learning respectively address the separation of samples among homogeneous and heterogeneous parties. To the best of our knowledge, though, a comprehensive framework that combines horizontal and vertical federated learning has not been developed. The problem setup has demonstrated, though, that it is likely for both types of data separation to be present; beyond the power system domain, different corporations also happen to possess data-holding branches in the same set of local districts, and such a framework is necessary for them to perform collaborative machine learning.

On the other hand, it is noteworthy that existing horizontal and vertical federated learning frameworks mostly choose to train XGBoost by transferring encrypted gradient statistics $G_{kv}$ and $H_{kv}$ to an ``active party", in order for this party to lead the training process. This consistent idea of gradient transfer within these frameworks allows us to formulate a hybrid federated learning framework based on XGBoost that merges both horizontal and vertical federated learning.

To this end, we propose a hybrid XGBoost-based federated learning framework. In \hyperref[Framework-2]{\textbf{Framework 2}}, we seamlessly integrate FederBoost \cite{FederBoost} with SecureBoost \cite{SecureBoost}, which are respectively a horizontal and vertical federated learning framework, to address both types of data separation. In this framework, we denote $C_m$ as the list of label-holding parties (i.e. transformer stations in the problem setup), and $D_m$ as the list of secondary parties (i.e. mobile stations in the problem setup). Each subscript $m$ denotes a particular district.

\begin{table}[ht]
\label{Framework-2}
\begin{tabular}{l  l}

\hline
\hline
\multicolumn{2}{l}{\textbf{Framework 2: Hybrid Federated Learning Based on XGBoost}} \\
\hline
\multicolumn{2}{l}{\textbf{Additional Notations:}} \\
\multicolumn{2}{l}{$C_m$, the label-holding party in district $m$ ($m=1,\cdots,M$)} \\
\multicolumn{2}{l}{$D_m$, the secondary party in district $m$ ($m=1,\cdots,M$)} \\
\multicolumn{2}{l}{$K_C,K_D$, number of features for the label-holding and secondary party} \\
\multicolumn{2}{l}{$G_{k,v,m}^{C/D}, H_{k,v,m}^{C/D}$, first-order/second-order gradient statistics on feature $k$,} \\ 
\multicolumn{2}{l}{\qquad \qquad bin $v$ and district $m$ of the label-holding/secondary party} \\
\multicolumn{2}{l}{$[[\cdot]], \text{Dec}\{\cdot\}$, Paillier encryption and decryption} \\ 
\ \\

\multicolumn{2}{l}{\textbf{District-Level Sample Alignment} \emph{prior to training}.} \\
\textbf{1} & \textbf{For} district $m = 1, 2, \cdots, M$: \\
\textbf{2} & \quad Parties $C_m$ and $D_m$ perform secure sample alignment \\
\textbf{3} & \quad Party $C_m$ encrypts all gradients of $[[g_i]]$ and $[[h_i]]$ \\
\textbf{4} & \textbf{End For}  \\

\ \\

\multicolumn{2}{l}{\textbf{Secure Gradient Transfer and Binning} \emph{at the start of each tree}.} \\
\textbf{1} & \textbf{For} district $m = 1, 2, \cdots, M$: \\
\textbf{2} & \quad $C_m$ calculates the first- and second-order gradients $g_i, h_i$ \\
\textbf{3} & \quad $C_m$ transfers Paillier-encrypted gradients $[[g_i]], [[h_i]]$ to $D_m$ \\
\textbf{4} & \textbf{End For} \\
\textbf{5} & Select an active district $m_A$ \\
\textbf{6} & \emph{Based on FederBoost's secure bin construction algorithm}, \\
\textbf{7} & Party $C_{m_A}$ leads the bin construction on features of every $C_m$\\
\textbf{8} & Party $D_{m_A}$ leads the bin construction on features of every $D_m$\\

\ \\

\multicolumn{2}{l}{\textbf{Secure Gradient Aggregation} \emph{for every node of a tree}.} \\
\textbf{1} & Select an active party $C_{m_B}$ from the label-holding parties \\
\textbf{2} & \textbf{For} district $m=1, 2, \cdots, M$: \\
\textbf{3}& \quad \emph{Based on \hyperref[Framework-1]{\textbf{Framework 1}}}, \\
\textbf{4} & \quad Party $C_m$ filters out samples not belonging to the current node \\
\textbf{5} & \quad Party $C_m$ calculates all gradient statistics $G_{k,v,m}^{C}$ and $H_{k,v,m}^{C}$ \\
\textbf{6} & \quad Party $C_m$ transfers $[[G_{k,v,m}^{C}]], [[H_{k,v,m}^{C}]]$ to $C_{m_B}$ \\
\textbf{7} & \quad Party $D_m$ filters out samples not belonging to the current node \\
\textbf{8} & \quad Party $D_m$ calculates encrypted statistics $[[G_{k,v,m}^{D}]], [[H_{k,v,m}^{D}]]$ \\
\textbf{9} & \quad Party $D_m$ transfers $[[G_{k,v,m}^{D}]]$ and $[[H_{k,v,m}^{D}]]$ to $C_{m_B}$ \\
\textbf{10} & \textbf{End For} \\
\textbf{11} & \makecell[l]{Party $C_{m_B}$ decrypts and sums gradient statistics \\ \ $G_{k,v}^{C}=\sum_m \text{Dec}\{[[G_{k,v,m}^{C}]]\}$, $H_{k,v}^{C}=\sum_m \text{Dec}\{[[H_{k,v,m}^{C}]]\}$} \\
\ \\
\multicolumn{2}{l}{\textbf{Node Splitting} \emph{for every non-leaf node of a tree}.} \\
\textbf{1} & Party $C_{m_B}$ calculates $G=\sum_{k}G_{k,1}^C$, $H=\sum_{k}H_{k,1}^C$ \\
\textbf{2} & \textbf{For} feature $k=1, 2, \cdots, K_C+K_D$: \\
\textbf{3} & \quad \textbf{For} bin  $v=1, 2, \cdots, V$: \\
\textbf{4} & \quad \quad Party $C_{m_B}$ calculates $\text{Gain}(k,v)$ \emph{based on \hyperref[Framework-1]{\textbf{Framework 1}}}  \\
\textbf{5} & \quad \textbf{End For} \\
\textbf{6} & \textbf{End For} \\

\textbf{7} & Party $C_{m_B}$ obtains $k^*$ and $v^*$ from the maximal $\text{Gain}(k,v)$ \\
\textbf{8} & If $k^*$ belongs to $C_{m_B}$, inform all $C_m$s, otherwise inform all $D_m$s \\
\textbf{9} & \makecell[l]{The parties holding $k^*$ (i.e., all $C_m$s or all $D_m$s) perform splitting \\ of node samples, and inform the other parties NOT holding $k^*$ \\ of the sample subspace of its child nodes} \\

\ \\
\multicolumn{2}{l}{\textbf{Leaf Value Calculation} \emph{for every leaf node of a tree}.} \\
\textbf{1} & Party $C_{m_B}$ calculates $G^C=\sum_{k}G_{k,1}^C$, $H^C=\sum_{k}H_{k,1}^C$ \\
\textbf{2} & Party $C_{m_B}$ sets leaf value \emph{based on \hyperref[Framework-1]{\textbf{Framework 1}}}  \\

\hline
\hline 

\end{tabular}
\end{table}

The framework contains four consecutive steps: 

\begin{enumerate}[wide]
\item \textbf{District-Level Sample Alignment}. For vertically-separated data in each district, samples need to be properly aligned before training. Secure alignment algorithms often need to be introduced not to leak the identity of individual samples; however, for the problem setup, sample IDs (i.e., the timestamps) contain no private information, and can therefore be aligned without encryption.

\item \textbf{Secure XGBoost Binning}. For horizontally-separated data across districts, the edges of bins in XGBoost are dependent on the statistic information of all homogeneous parties. Therefore, we resort to FederBoost's secure bin construction algorithm \cite{FederBoost} to find proper bin edges. For the label-holding parties, the active party $C_{m_A}$ leads the bin construction process on features of all $C_m$s; the same goes for secondary parties.

\item \textbf{Secure Gradient Aggregation}. For a target node to be trained, each party filters out samples not belonging to the target node from the bins. Then each party calculates the gradient statistics of each bin, and securely transfer the statistics to the label-holding active party $C_{m_B}$. The active parties, upon receiving the encrypted gradients, sum and decrypt the statistics for each bin and feature.

\item \textbf{Node Splitting or Leaf Value Calculation}. For non-leaf nodes, the active party $C_{m_B}$ calculates $\text{Gain}(k,v)$ iteratively for every feature $k$ and bin edge $v$ based on the aggregated gradient statistics, and obtains the optimal $(k^*, v^*)$ with the maximal $\text{Gain}$. Then the active party informs the splitting results to the  parties ($C_m$s or $D_m$s) that hold the feature $k^*$, who will perform node splitting, determine the sample subspace of child nodes, and inform the parties not holding $k^*$ of the subspace. On the other hand, if the target node is a leaf node waiting for a value to be set, then only the label-holding active party $C_{m_B}$ needs to perform calculation based on the aggregated gradient statistics. 

\end{enumerate}

\begin{figure}
\centering
\includegraphics[scale=0.33]{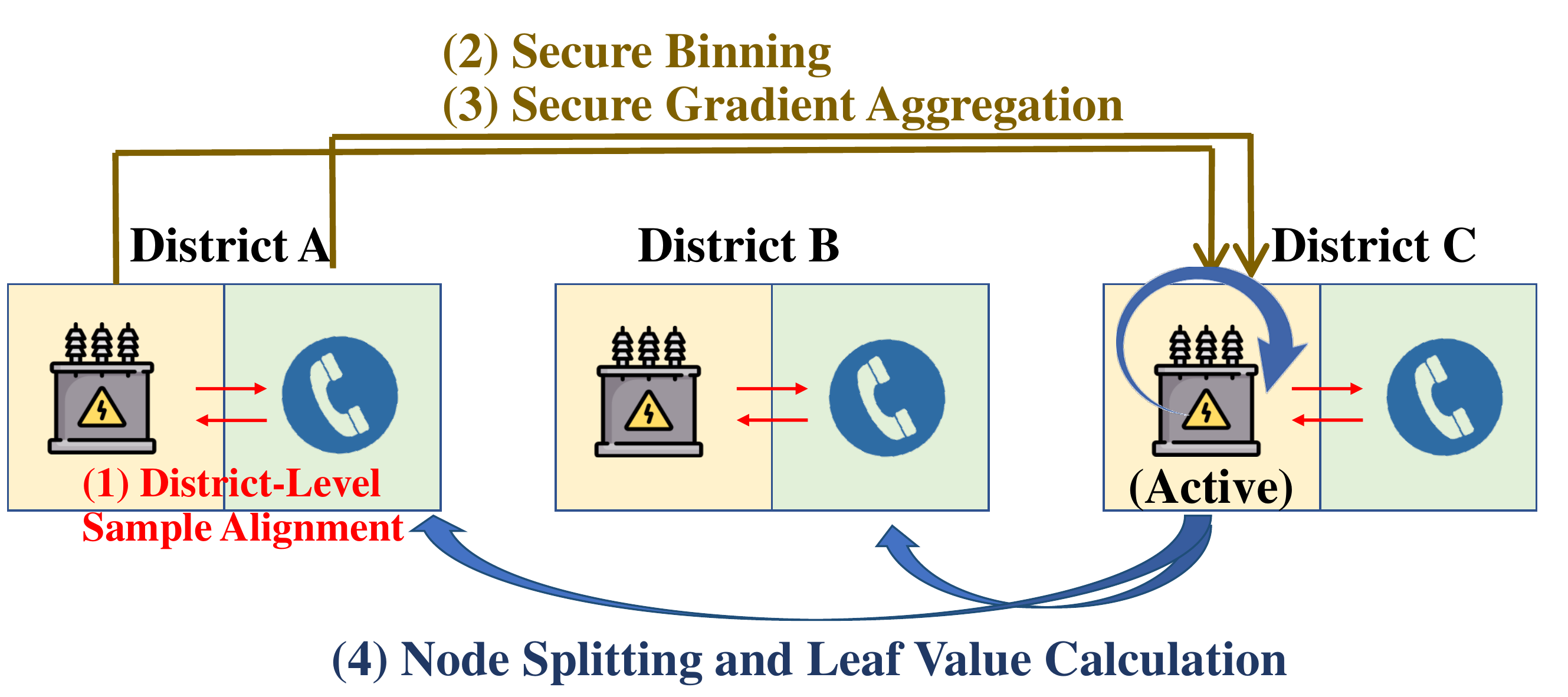}
\caption{Application of hybrid federated learning (\hyperref[Framework-2]{\textbf{Framework 2}}) in the given problem setup.}
\label{Framework-2-Fig}
\end{figure}

\hyperref[Framework-2-Fig]{\textbf{Figure 2}} demonstrates the application of \hyperref[Framework-2]{\textbf{Framework 2}} in the given problem setup. Since the transformer stations hold the power consumption labels, one of the transformer stations is expected to become the active party and perform the node splitting task for each training epoch. All other transformer stations and mobile branches are obliged to send the bin statistics and aggregated gradients in steps 2) and 3). By leveraging both vertical SecureBoost and horizontal FederBoost, a collaborative XGBoost model is now trained with participation from all districts and both heterogeneous parties (namely the transformer stations and China Mobile). The resulting model is therefore expected to have a higher prediction accuracy and generalization ability with the additional samples and features supplemented into the dataset.

\subsection{Dynamic Task Allocation Scheme}
In vertical SecureBoost and horizontal FederBoost, the two building blocks of our hybrid federated learning framework, the active party (i.e., $C_{m_B}$) is fixed throughout the training process. This raises fairness and efficiency issues, as (a) The active party receives the whole intermediate information, while the other homogeneous parties, despite an equal contribution of training data, do not. This gives the active party an unfair advantage at inferring data of other parties; (b) The gradient aggregation and node-splitting tasks are all processed by the active party's computing resources only, while other parties passively wait for the results, which is time-consuming. To this end, we propose a dynamic task allocation scheme in \hyperref[Framework-3]{\textbf{Framework 3}}.

\begin{table}[ht]
\label{Framework-3}
\begin{tabular}{l  l}

\hline
\hline
\multicolumn{2}{l}{\textbf{Framework 3: Dynamic Task Allocation}} \\
\hline

\multicolumn{2}{l}{\textbf{In each training round of an epoch:}} \\
\textbf{1} & \makecell[l]{Select a node from the queue with the smallest index for training.} \\

\textbf{2} & \makecell[l]{Select the currently most available party as the active party.} \\

\textbf{3} & \makecell[l]{The active party initiates gradient aggregation and node-splitting.} \\

\textbf{4} & \makecell[l]{Add the left and right child nodes of the current node to the queue.} \\

\hline
\hline 

\end{tabular}
\end{table}

\begin{figure}
\centering
\includegraphics[scale=0.3]{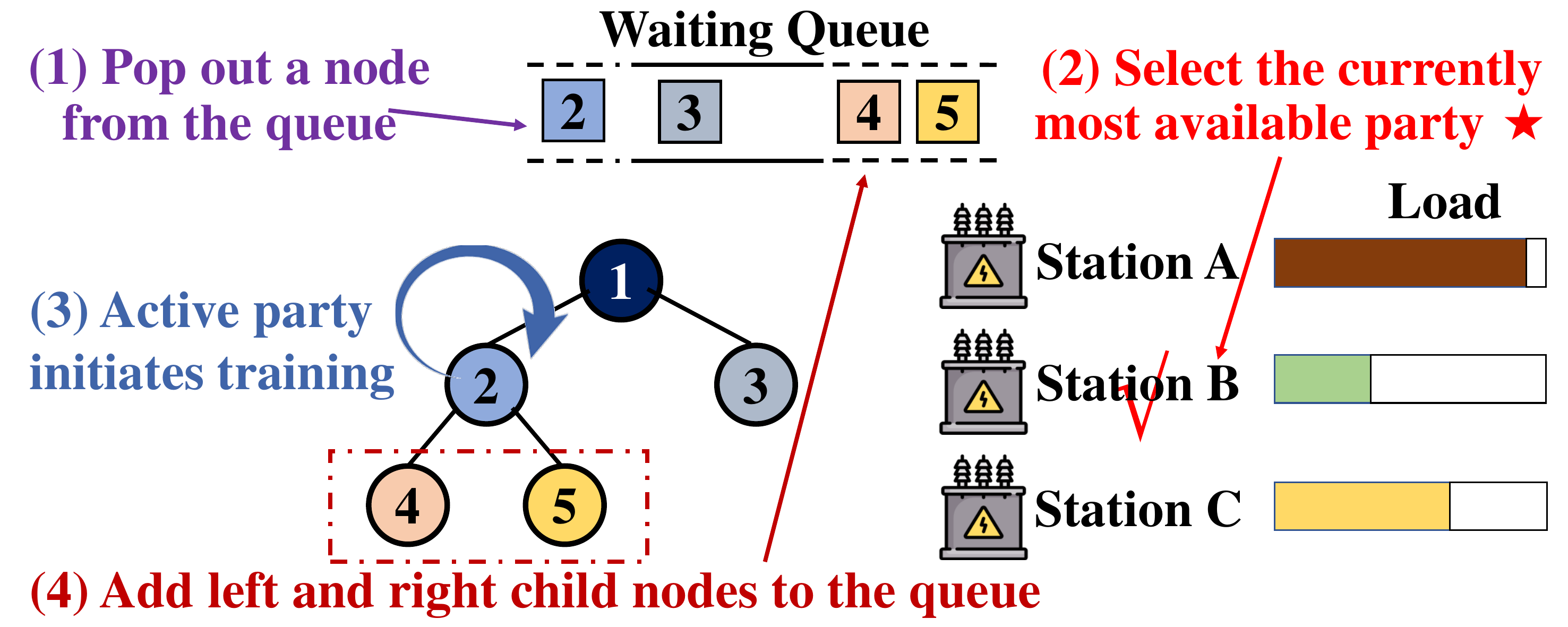}
\caption{Application of the dynamic task allocation scheme (\hyperref[Framework-3]{\textbf{Framework 3}}) in the given problem setup.}
\label{Framework-3-Fig}
\end{figure}

The underlying mechanism of the framework is intuitive. Every time we select a node with the smallest index from the waiting queue, and assigns the currently most available party as the active party. After the active party organizes and finishes training on the current node, it adds the left and right child node of this node into the queue. Also, the computational resources of all parties can be fully leveraged with alternating active parties. With the dynamic assignment of active parties, all parties have fairer access to the intermediate gradient statistics. The application of \hyperref[Framework-3]{\textbf{Framework 3}} in the problem setup is plotted in \hyperref[Framework-3-Fig]{\textbf{Figure 3}}, in which the selected transformer station, namely Station B due to the lightest computational load, coordinates the node-splitting task of the target node.

\section{Performance Evaluation}
\label{Section-IV}
In this section, we theoretically investigate the performance of the proposed framework. Specifically, we confirm the security of the framework, and demonstrate its superiority in terms of fairness and efficiency. Another important property of the framework, losslessness against a centralized and unencrypted XGBoost, is axiomatic, as long as the precision before encrypting floating-point gradients is set high enough (e.g., round gradients to a scale of $10^{-6}$). We therefore do not elaborate on this property as a separate section.

\subsection{Security}
The security of the framework is guaranteed based on the following observations. First, all individual raw data are stored locally, with only timestamp IDs being publicly transferred. Second, all intermediate variables for training (e.g., individual gradients and gradient statistics) are Paillier-encrypted before being transferred. We can easily balance the trade-off between encryption quality and efficiency by tuning the magnitude of the prime number in Paillier. Third, each party has access to the node-splitting criteria if and only if it holds the feature that splits that node. Even after training, each party only holds half of the XGBoost model, and a collaborative learning is required in order to predict labels with the model \cite{SecureBoost}.

We further note that with a dynamic task allocation scheme, the gradient statistics are now transferred to different active parties when splitting all the non-leaf nodes. This gives all parties a smaller chance at inferring the properties of any individual data with only part of the gradients.

\subsection{Fairness}
\label{fairness-subsection}
Fairness measures are often used in network engineering to determine whether users are receiving a fair share of system resources. Under the construct of hybrid federated learning, they can be extended to study the fairness of parties in receiving intermediate information, which can lead to easier inference of other parties' properties. Specifically, we employ Jain's fairness index $\mathcal{J}$ \cite{Jain}:

\begin{equation}
\mathcal{J}(x_1,x_2,\cdots, x_M)=\frac{(\sum_{m=1}^{M}x_m)^2}{M \cdot \sum_{m=1}^{M}x_m^2}
\end{equation}

\noindent where $x_1,x_2,\cdots, x_M$ denote the amount of information received by each party. $\mathcal{J}$ takes value between $1/M$ and 1, and a larger $\mathcal{J}$ indicates greater fairness.

With the dynamic task allocation scheme, the fairness of the framework can be improved. To demonstrate, we first make the ideal assumption of a uniform computational load: the gradient aggregation time $\tau_1^m$ and node-splitting time $\tau_2^m$ for each party $m$ to process each node are fixed to $\tau_{10}$ and $\tau_{20}$ respectively. All communicational time between parties has been reduced to either the gradient aggregation side or the node-splitting side. 

Under this assumption, in face of task conflicts assigned to a same party, we should take priority on tasks that come in first, as this can result in the highest computational efficiency. The party assigned as the active party of each non-leaf node can therefore be pre-determined. Let $l$ be the layer index of the current node and $n$ be the total number of non-leaf layers, then we can prove the task allocation scheme to be:

\begin{enumerate}
\item Nodes from layer 1 to layer $\lceil log_2 M \rceil$ are assigned to parties $1,2, \cdots, 2^{l-1}$ to perform node-splitting as the active party.

\item Nodes from layer $\lceil log_2 M \rceil + 1$ to layer $n$ are assigned to parties $1,2, \cdots, M$ repeatedly in a breadth-first manner, to perform node-splitting as the active party.

\end{enumerate}

The fairness index can then be deduced as

\begin{equation}
\label{fair-eq}
\mathcal{J}=
\begin{cases}
\frac{(2^n-1)^2}{M(3 \times 2^n-2n-3)},& M > 2^{n-1} \\
\frac{(2^n-1)^2}{\sum_{m=1}^M(\lceil \text{log}_2M\rceil - \lceil \text{log}_2m \rceil +p+\alpha(M))^2},& M\leq 2^{n-1}
\end{cases}
\end{equation}

\noindent where $\alpha(M)= \begin{cases} 1, & 1 \leq m \leq q \\ 0, & q+1 \leq m \leq M \end{cases}$. $p$ and $q$ are the quotient and remainder of $2^n-2^{\lceil \text{log}_2M \rceil}$ divided by $M$. 

\begin{figure}
\centering
\includegraphics[scale=0.35]{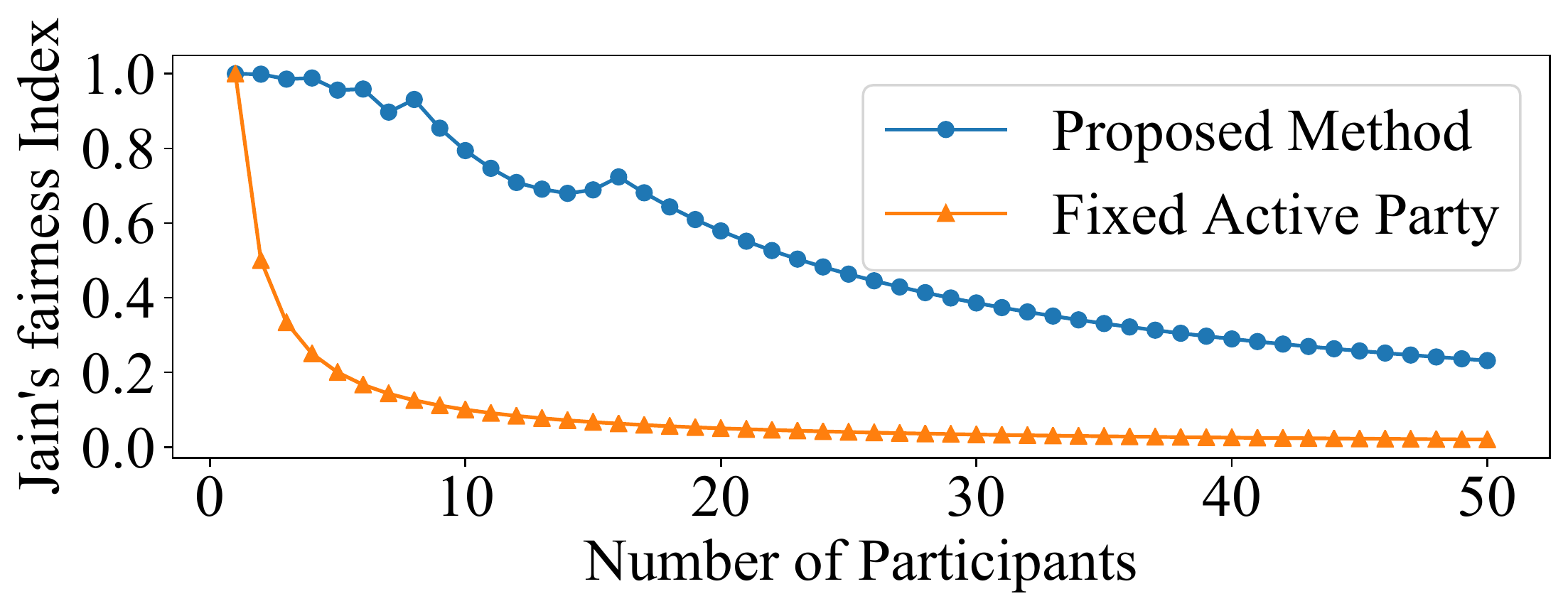}
\caption{Jain's fairness index of the hybrid federated learning framework, under a dynamic task allocation scheme (blue) and a fixed task allocation scheme (orange). $n$ is set to 5.}
\label{Fairness}
\end{figure}

The Jain's fairness index of the hybrid federated framework under a dynamic task allocation scheme is compared with that under a fixed active party scheme in \hyperref[Fairness]{\textbf{Figure 4}}. It can be seen that, as the number of active parties $M$ increases, both fairness curves drop proportionally to $1/M$, but their coefficients differ by a factor of $2^N / 3M$, indicating much higher fairness for all $M $s. More importantly, when $M<2^{n-1}$, the fairness of the proposed scheme drops significantly slower as $M$ increases (hardly affected when $M \leq 5$). 

Intuitively in non-ideal cases, the choice of active parties cannot be pre-determined, hence there is no general formula for fairness. However, when all the $\tau_1^{m}$s and $\tau_2^{m}$s are sufficiently uniform, the node-splitting tasks can still be quasi-uniformly allocated to each party, and that the resulting fairness is still outstanding. In the most extreme cases when $\tau_1^m$s and $\tau_2^m$s are statistically different to each other, fairness will be somewhat compromised, as node-splitting tasks will be inclined towards parties with more computational power, but usually still much higher than that of the fixed active party scheme. We will investigate into this effect in \hyperref[Section-V]{\textbf{Section V}}. 

\subsection{Efficiency}
With the proposed task allocation scheme, the node-splitting tasks are dynamically allocated to $M$ parties, mathematically identical to as if there were a centralized computing device with $M$ parallel pools. Parallel computing can therefore be fully leveraged to accelerate training. In practical scenarios, the number of parties $M$ is often much larger than the number of parallel pools in a single computer, so that computational efficiency can be significantly higher.

To illustrate, we still base our analysis on the ideal assumption and conflict management scheme introduced in the previous subsection. The total computational time $T_M$ can then be deduced as

\begin{equation}
T_M  = \tau_{10} (2^n-1) + \tau_{20} \{ \lceil \text{log}_2 M \rceil  + \lceil \frac{1}{M} (2^n-2^{ \lceil \text{log}_2M \rceil})\rceil \}
\end{equation}

\noindent which decreases as the total number of parties $M$ increases.

\begin{figure}
\centering
\includegraphics[scale=0.35]{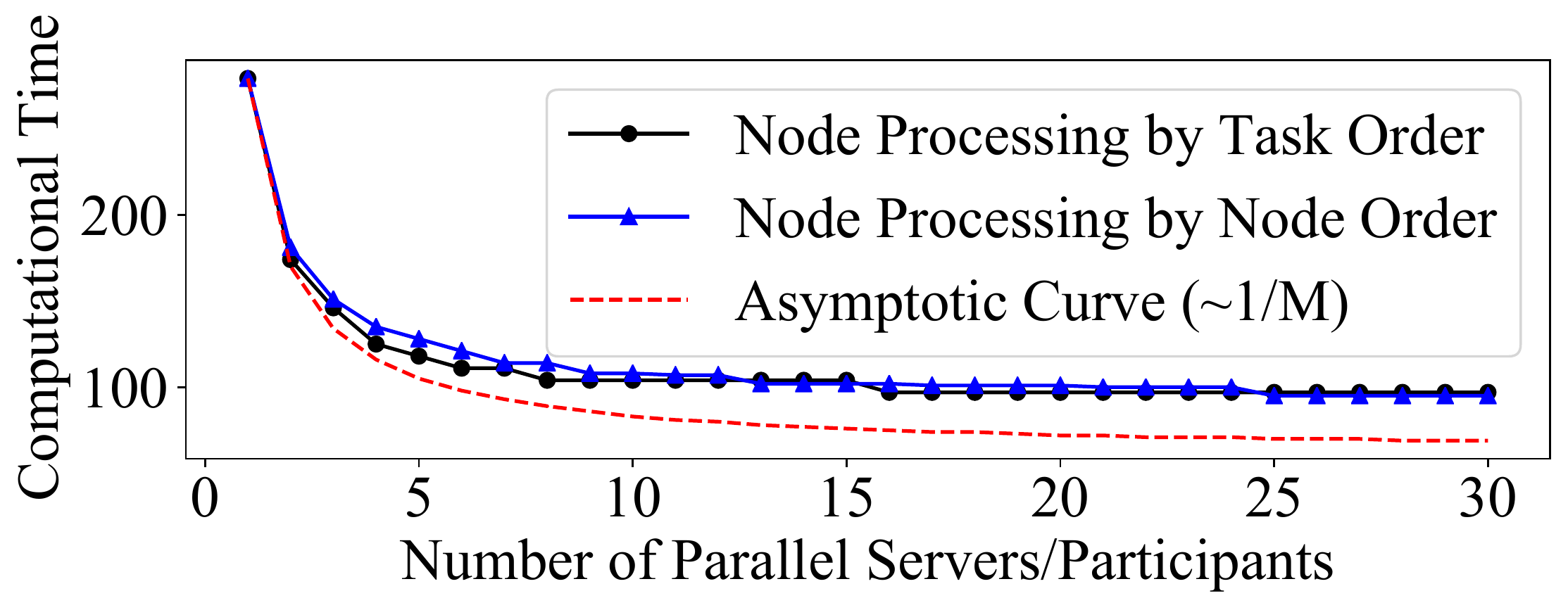}
\caption{Computational time following task order (optimal) and breadth-first node order (non-optimal) in face of task conflicts. $n=5$, $\tau_{10}=2$ and $\tau_{20}=7$.}
\label{efficiency}
\end{figure}

\hyperref[efficiency]{\textbf{Figure 5}} simulates the computational time required for training, under $n=5$, $\tau_{10}=2$ and $\tau_{20}=7$. Node processing by task order in face of task conflicts is proven to be slightly more efficient than that by breadth-first node order, which justifies our choice of the conflict management scheme. Also, the computational time follows an asymptotic decay of $~1/M$ when $n$ is sufficiently large:

\begin{equation}
T_M / T_\infty \rightarrow 1 + \frac{\tau_{20}}{M \tau_{10}}.
\end{equation}

Hence, the marginal efficiency improvement is most significant when $M$ is low. Recalling that the fairness among parties also drops most slowly at the lower end of $M$, we can conclude that the marginal benefits of the proposed dynamic allocation schemes are most outstanding for a small number of participating parties.

Similar to fairness, it is unlikely to derive an analytical expression for efficiency when the computation time is non-ideal. However, the computational time is still expected to be much higher than a fixed active party scheme. In a non-ideal situation where the computational time is statistically different among parties (See \hyperref[Section-V]{\textbf{Section V}} for case study), the total computational time is approximately the time required for the slowest party to finish gradient aggregation, as all the node-splitting tasks will be dynamically allocated to faster parties.

\section{Case Study}
\label{Section-V}
In this section, we perform a comprehensive case study on the problem setup. We implement the proposed XGBoost-based hybrid federated learning framework to predict distributed real-time power consumption of 10 spatially-adjacent districts, which span over a longitude of 0.1$^\circ$ and latitude of 0.06$^\circ$. The training set spans over a time range of 2019.1.1 -- 2020.1.20, and the test set spans over a time range of 2020.1.23 -- 2020.3.21. All labels, namely the real-time power consumption, are normalized before training. Also, the dynamic task allocation scheme is implemented to improve fairness and efficiency. The case study is implemented in C++ (MinGW32-GCC 4.8.2) for constructing a modified version of XGBoost, and Python 3.8 for fairness and efficiency evaluation.

To demonstrate the superiority of XGBoost as a bottom-level machine learning algorithm, in \hyperref[bottom-level]{\textbf{Table 2}} we compare the prediction accuracy of different learning algorithms after parameter tuning, in terms of the MSE. It can be seen that although the last three algorithms achieve a low prediction error on the training set, artificial neural network is prone to overfitting and sometimes even fails to converge. On the other hand, tree-based networks such as random forest and XGBoost generalize better to the test set, and XGBoost is proven to be the most outstanding. It should be noted that time-serie power forecasting based on historic consumption records reaches a test set MSE of 0.1788, indicating our prediction method based on real-time external features is sufficiently accurate.

\begin{table}[ht]
\label{bottom-level}
\begin{center}

\caption{Mean squared error (MSE) of bottom-level machine learning algorithms on predicting power consumption.}

\begin{tabular}{c  c  c  }
\hline
\hline
\textbf{Algorithm} & \textbf{Training Set MSE} & \textbf{Test Set MSE} \\

\hline
Dummy Regressor & 0.8751 & 0.9449 \\
Artificial Neural Network & 0.1491 & 0.2954 \\
Random Forest & 0.0874 & 0.1072 \\
XGBoost & 0.0528 & 0.0948 \\

\hline
\hline 
 
\end{tabular}
\end{center}
\end{table}

XGBoost is then integrated into the hybrid federated learning framework as outlined in \hyperref[Section-III]{\textbf{Section III}}. To demonstrate its performance in prediction, we construct 4 different federated learning settings for comparison.

Case 1: Only horizontal federated learning is adopted. All transformer stations collaboratively train an XGBoost model without participation from China Mobile.

Case 2: Only vertical federated learning is adopted. A transformer station only collaborates with the mobile station in the same district to train an XGBoost model. (The test set is also restricted to data from this particular district.)

Case 3: Hybrid federated learning but without encryption.

Case 4: Hybrid federated learning with Paillier-based homomorphic encryption.

\begin{figure}
\centering
\includegraphics[scale=0.45]{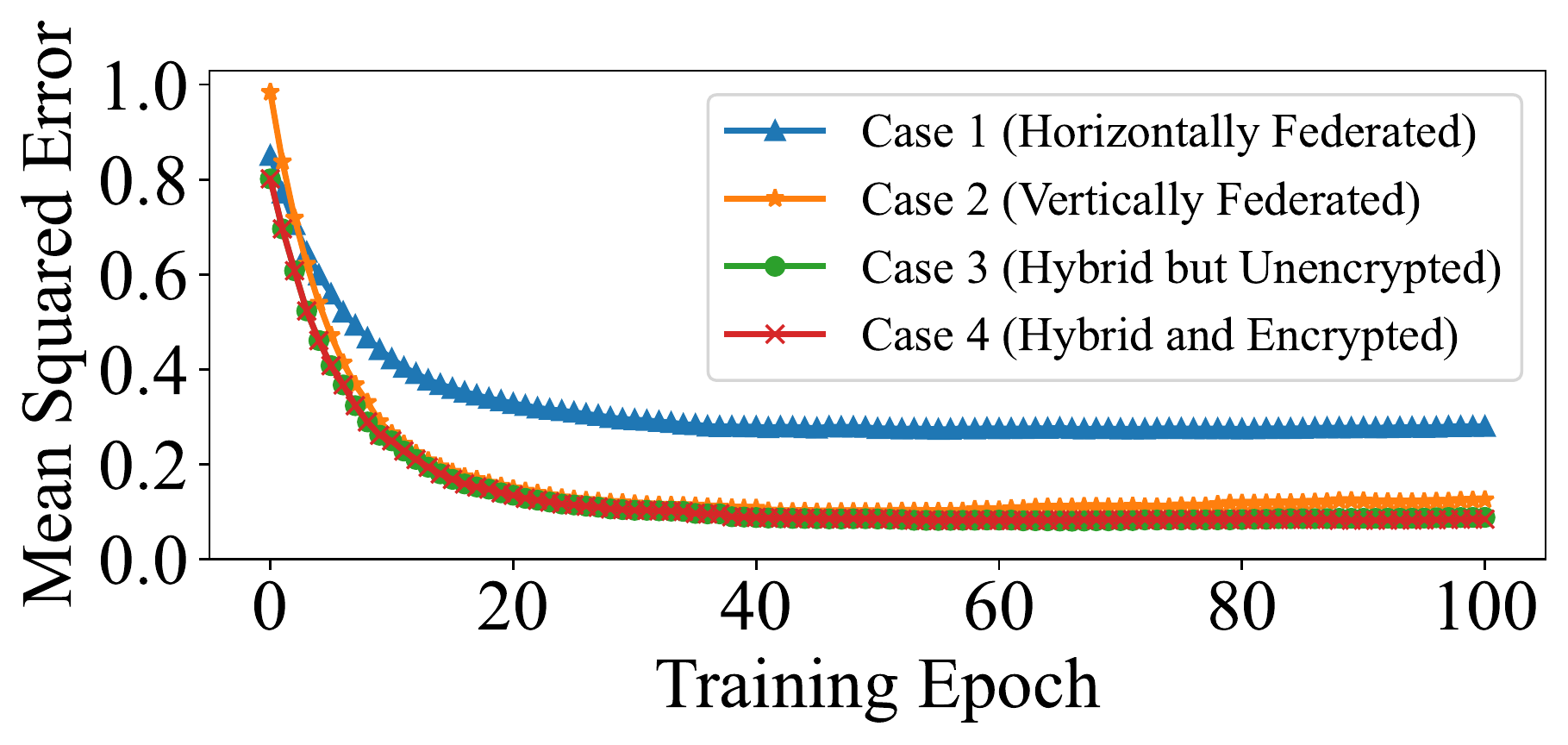}
\caption{Prediction MSE on the test set, for the 4 cases of federated learning.}
\label{Convergence}
\end{figure}

\hyperref[Convergence]{\textbf{Figure 6}} compares their prediction MSE on the test set during each training epoch. It is evident that the prediction accuracy of Case 1 is significantly degraded due to the absence of China Mobile. Cases 2, 3 and 4 reach a similarly MSE of $0.091 \pm 0.007$ after training, with almost overlapping training curves, which suggests that (a) there is little accuracy loss in prediction (and even accuracy gain in some districts) after horizontally incorporating the data of other districts; (b) the integrated Paillier encryption scheme is lossless.

\begin{figure}
\centering
\includegraphics[scale=0.4]{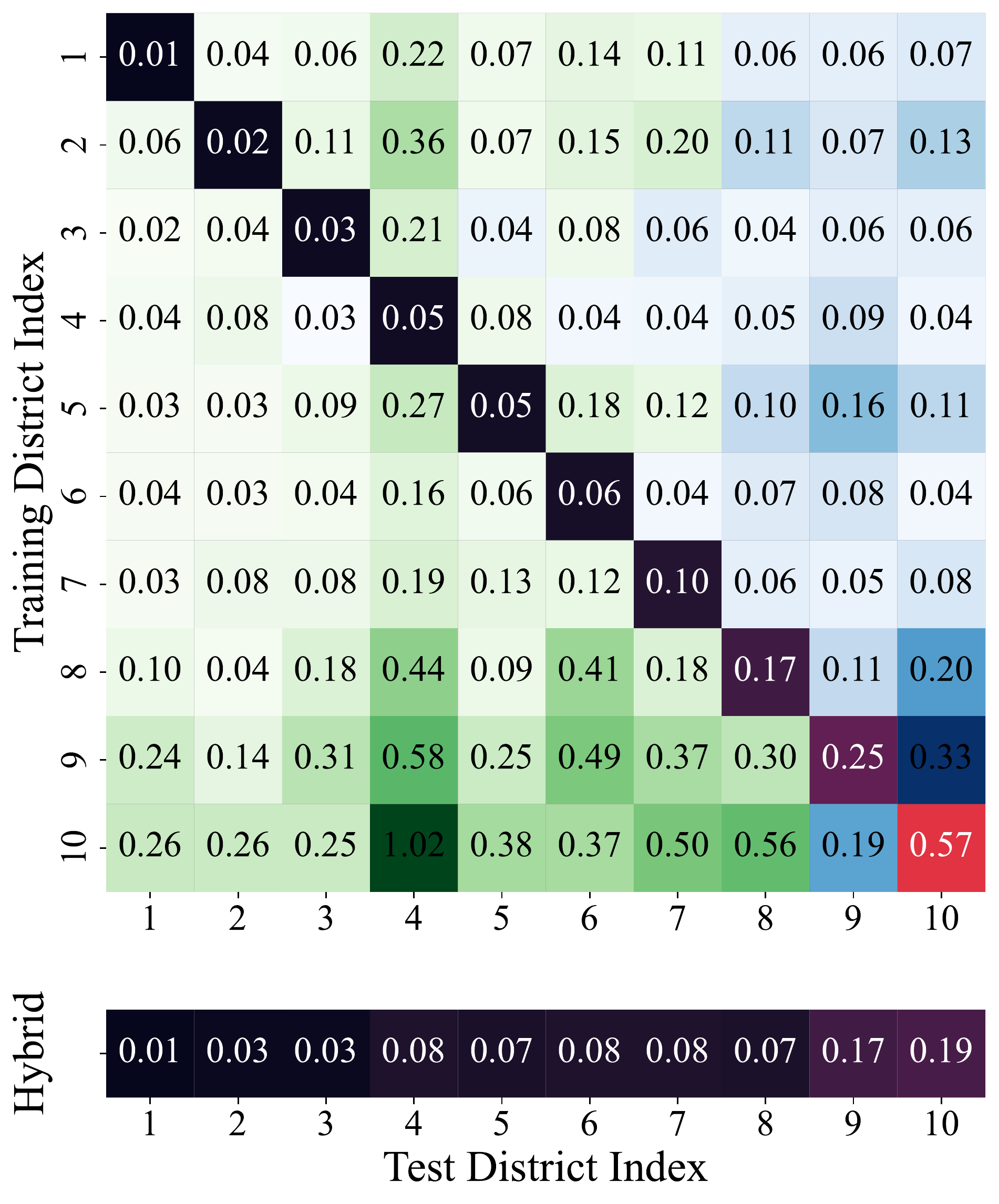}
\caption{\textbf{Up}: Prediction MSE with models vertically trained and tested respectively on each single district. Green blocks indicate scenarios where the vertically-trained model predicts worse than one trained on data of the test district (i.e., the diagonal blocks of the same column). Blue blocks indicate otherwise. \textbf{Down}: Prediction MSE of the hybrid learning model tested on each district.}
\label{Heatmap}
\end{figure}

Furthermore, we observe that models trained by hybrid federated learning in fact have a higher generalization ability on all districts under study. \hyperref[Heatmap]{\textbf{Figure 7}} plots an adjusted heatmap of the MSE error vertically trained and tested respectively on each of the 10 single districts, as well as a heatmap of the prediction MSE of the hybrid learning model on each district. Blocks with a green colormap indicate scenarios where the as-trained model predicts worse on the test district than a model trained exactly on the training data of the same district (i.e., the diagonal blocks of the same column), and blocks with a blue colormap indicate otherwise. It can be seen that the vertically-trained single-district models mostly generalize poorly to other districts. The only exceptions are Districts 8-10 where the models of other districts may seem to generalize surprisingly well. However, this specious intuition can be debunked by the fact that these districts have much higher diagonal MSEs (a maximum of 0.57) compared to others, indicating that their training datasets are unrepresentative even of their districts, leading to inappropriate MSE baselines for comparison. Regardless, hybrid federated learning manages to predict accurately on each of the 10 districts with an enriched training dataset, with a maximum MSE of only 0.19.

\begin{figure}
\centering
\includegraphics[scale=0.45]{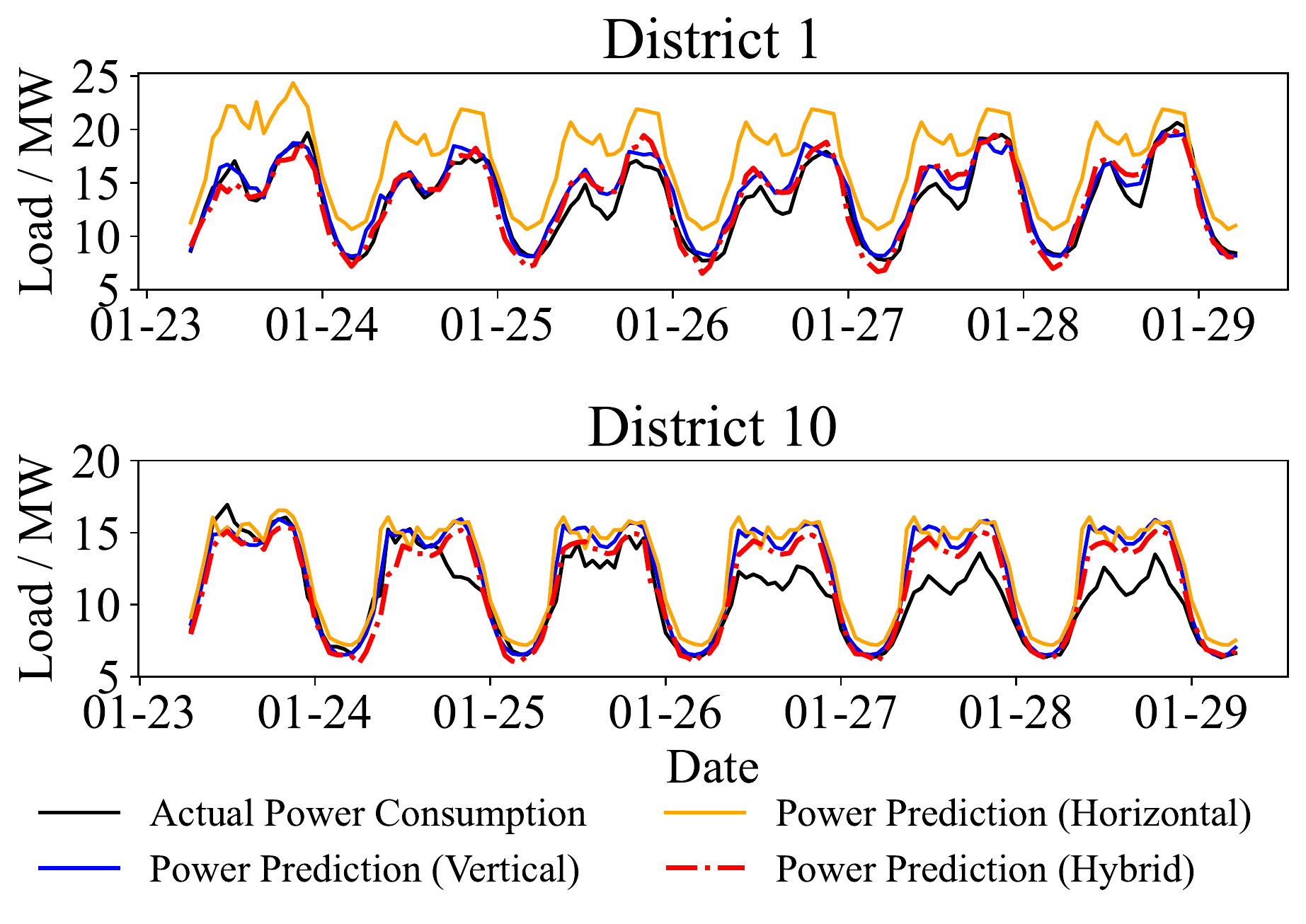}
\caption{Actual power curve over a range of 6 days on Districts 1 (\textbf{Up}) and 10 (\textbf{Down}), as well as the corresponding predicted power curves with vertical, horizontal and hybrid federated learning.}
\label{Power-Curve}
\end{figure}

We also validate the real-time power prediction results by visualizing the power curves on the test set of Districts 1 and 10 (\hyperref[Power-Curve]{\textbf{Figure 8}}), which according to the heatmaps have the lowest and highest prediction error. In District 1, vertical and hybrid federated learning frameworks predict equally well on the test set, with the predicted power curves almost overlapping with the actual one. Horizontal federated learning fails to reach a comparable performance due to the absence of features from China Mobile. Interestingly in District 10, the actual power curve experiences a steady downturn after January 25, attributable to the week-long Lunar New Year's festival in China and the outbreak of Covid-19. All three federated learning frameworks fail to predict accurately in this particular scenario, but hybrid federated learning, possibly due to incorporation of the most datasets from other districts, generates a predicted curve closest to the actual curve.

\begin{figure}
\centering
\includegraphics[scale=0.5]{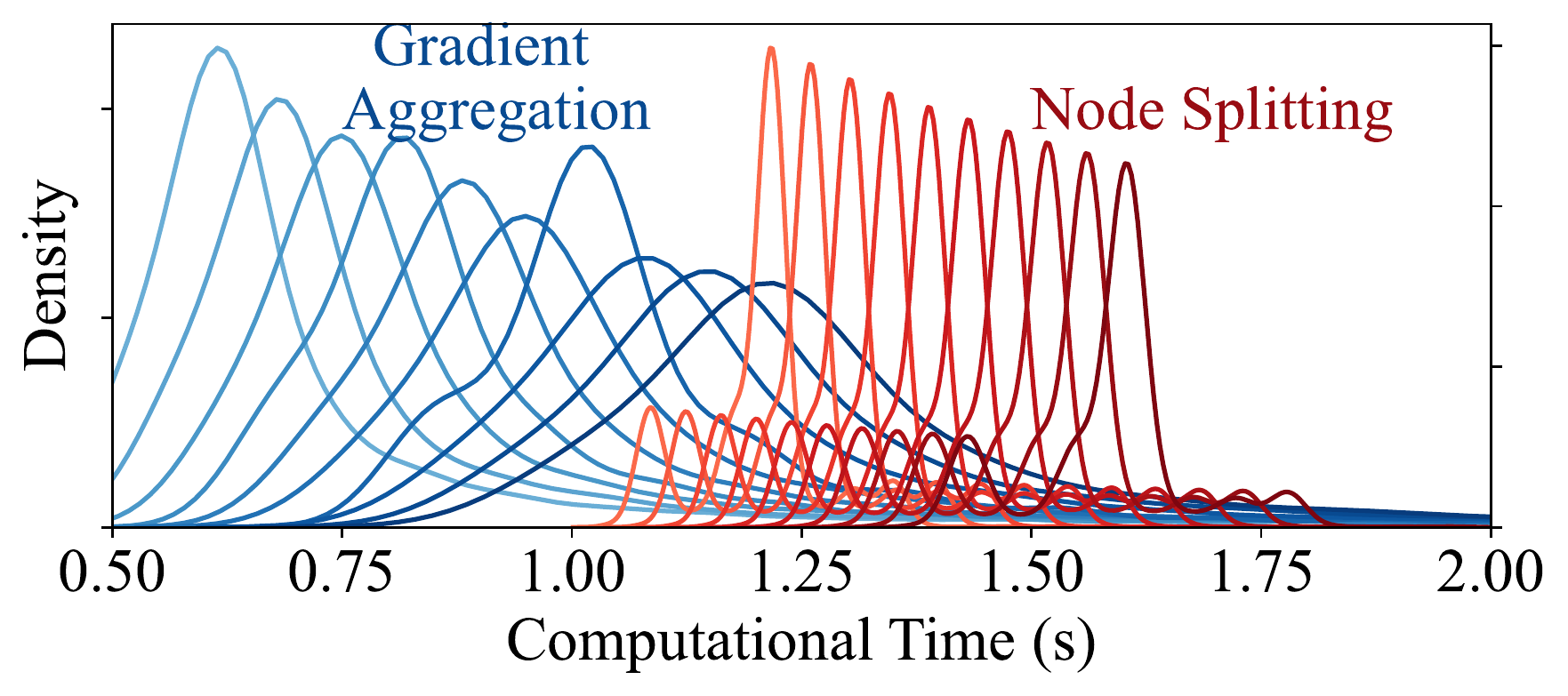}
\caption{Kernel density of the distribution of computational time in the 10 districts, for the gradient aggregation step (Blue) as well as the node-splitting step (Red) on each node of XGBoost.}
\label{Density}
\end{figure}

Finally, we simulate and evaluate the efficiency and fairness of the hybrid federated learning framework with the dynamic task allocation scheme. Different from the ideal assumption made in \hyperref[Section-IV]{\textbf{Section IV}}, the computation time is now statistically different among different districts (See \hyperref[Density]{\textbf{Figure 9}}), with a maximum p-value of 0.0067 for gradient aggregation and $8 \times 10^{-163}$ for node-splitting in two-sided T-tests.

\begin{figure}
\centering
\includegraphics[scale=0.5]{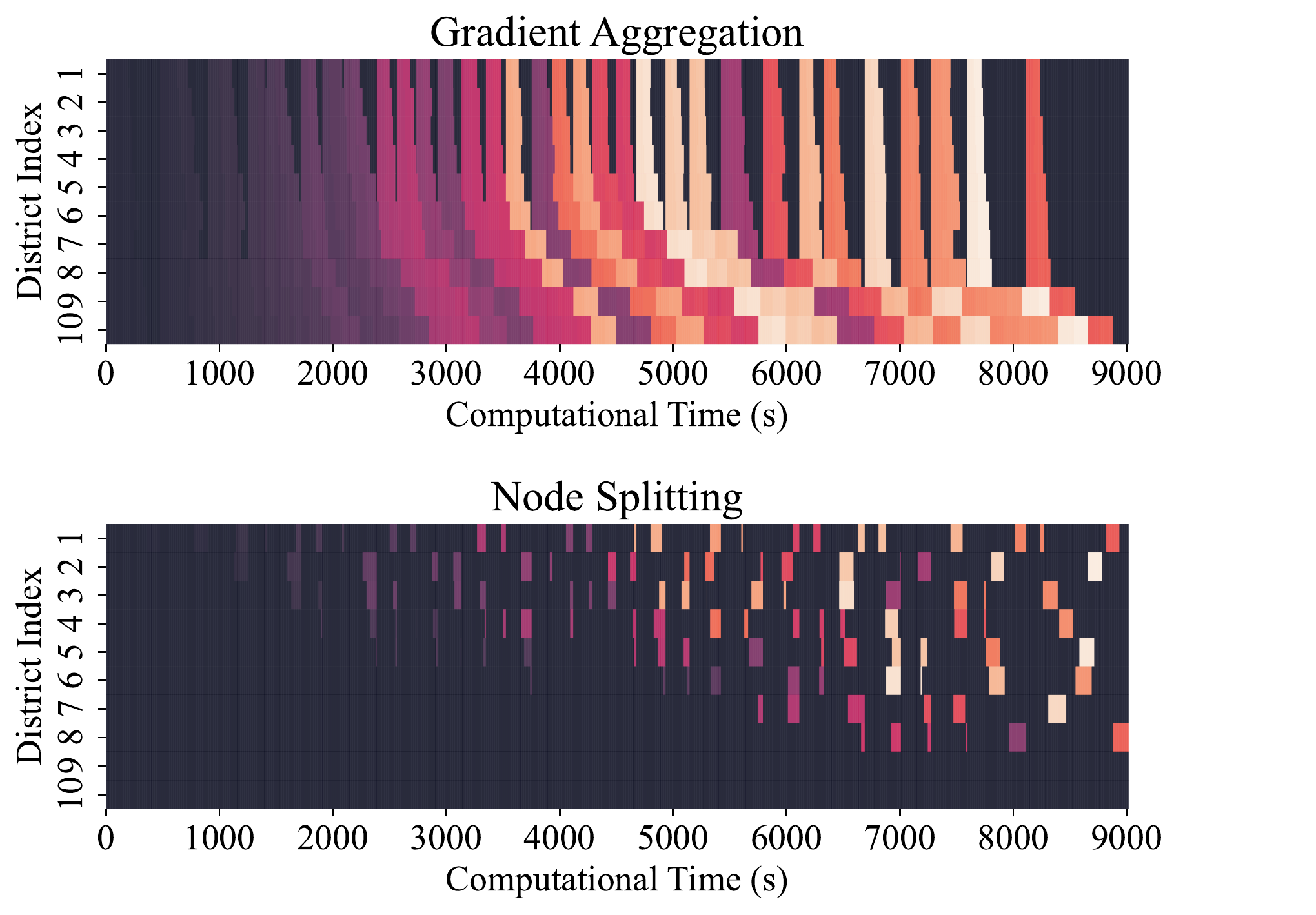}
\caption{Allocation of computational load during the first epoch of training, in terms of gradient aggregation (\textbf{Up}) and node-splitting (\textbf{Down}), with the dynamic task allocation scheme. Colors indicate indices of the XGBoost nodes currently under calculation. Lighter colors indicate larger node indices. Black blocks indicate idle time.}
\label{Computational-Load}
\end{figure}

\hyperref[Computational-Load]{\textbf{Figure 10}} plots the computational load allocated among the 10 districts during the first epoch of training. With the dynamic task allocation scheme, the training now requires only 9,017 seconds to complete, and may otherwise require an additional 7,220 seconds if we adopted a fixed active party allocation scheme instead. A closer investigation suggests that with the dynamic task allocation scheme, the node-splitting tasks are now more inclined towards districts with higher computational power; in fact, Districts 9 and 10 shoulder no node-splitting tasks due to their computational slowness. This further boosts computational efficiency, as the total gradient aggregation time is now identical to the least time required, namely the time for the slowest party to finish its own aggregation.

\begin{figure}
\centering
\includegraphics[scale=0.4]{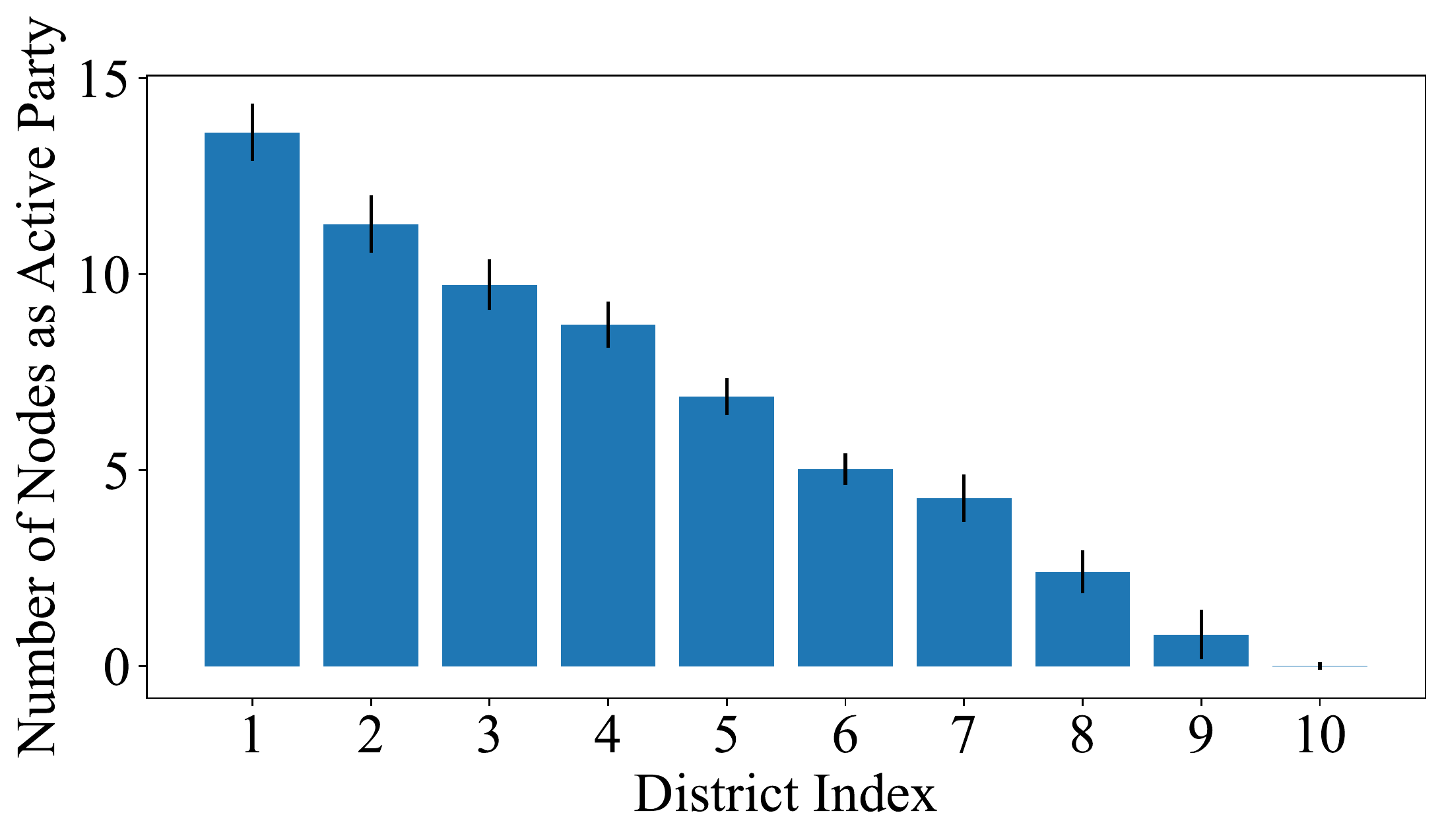}
\caption{Average number of nodes that each district is assigned as the active party in the hybrid federated learning process. Error bars indicate standard deviation across epochs.}
\label{Fairness-Case}
\end{figure}

In terms of fairness, \hyperref[Fairness-Case]{\textbf{Figure 11}} plots the average number of nodes across epochs that each district is assigned as the active party. It is evident that the probability of active party assignment decreases as the computational power decreases, in exchange for higher computational efficiency. The resulting fairness index $\mathcal{J}$ therefore lies in the range of [0.609,0.739], which is somewhat smaller than the ideal situation of 0.947 but much larger than the fixed active party scheme with a fairness of 0.1. This suggests that there is a trade-off between fairness and efficiency in the dynamic task allocation scheme.

\section{Conclusion}
\label{Section-VI}
In this paper, we propose a hybrid federated learning framework that combines both horizontal and vertical federated learning based on XGBoost, for distributed power prediction of a city in southern China. With the hybrid framework, homogeneous and heterogeneous parties can collaboratively train an XGBoost model without leaking private information. In addition, a dynamic task allocation scheme is proposed to improve training performance. The follow-up case study justifies the practicability of the proposed framework, and its in providing a secure, fair and computationally efficient solution for the power prediction problem.

The hybrid federated learning framework with the dynamic task allocation scheme have deep implications in the power system. It addresses the distributed nature of smart grid data in both the horizontal and vertical dimension, and guides all parties to contribute to the pattern mining of power generation/consumption data in a coordinated manner. This allows for timely and effective decision-making after the power prediction process. Other potential improvements to this framework include the consideration of communication delays to address system heterogeneity, as well as the adoption of multi-task learning to address statistical heterogeneity.

\end{document}